\documentclass{article}
\usepackage[T1]{fontenc}
\usepackage[utf8]{inputenc}
\usepackage{spconf,amsmath,graphicx,hyperref}
\usepackage{booktabs}
\usepackage{multirow}
\usepackage{amsfonts}

\title{Temporal-Spatial Decouple before Act: Disentangled Representation Learning for Multimodal Sentiment Analysis}

\name{Chunlei Meng$^{1}$, Ziyang Zhou$^{2}$, Lucas He$^{3}$, Xiaojing Du$^{4}$, Chun Ouyang$^{1}$\,$^{\dagger}$, Zhongxue Gan$^{1}$\,\thanks{$^{\dagger}$Corresponding author.}}
\address{$^{1}$Fudan University, $^{2}$Shantou University, $^{3}$University College London, $^{4}$University of South Australia}

\begin{document}
\ninept
\maketitle

\begin{abstract}
Multimodal Sentiment Analysis integrates Linguistic, Visual, and Acoustic. Mainstream approaches based on modality-invariant and modality-specific factorization or on complex fusion still rely on spatiotemporal mixed modeling. This ignores spatiotemporal heterogeneity, leading to spatiotemporal information asymmetry and thus limited performance. Hence, we propose TSDA, Temporal–Spatial Decouple before Act, which explicitly decouples each modality into temporal dynamics and spatial structural context before any interaction. For every modality, a temporal encoder and a spatial encoder project signals into separate temporal and spatial body. Factor-Consistent Cross-Modal Alignment then aligns temporal features only with their temporal counterparts across modalities, and spatial features only with their spatial counterparts. Factor specific supervision and decorrelation regularization reduce cross factor leakage while preserving complementarity. A Gated Recouple module subsequently recouples the aligned streams for task. Extensive experiments show that TSDA outperforms baselines. Ablation analysis studies confirm the necessity and interpretability of the design.
\end{abstract}


\begin{figure*}[htbp]
    \centering
    \includegraphics[width=0.9\linewidth]{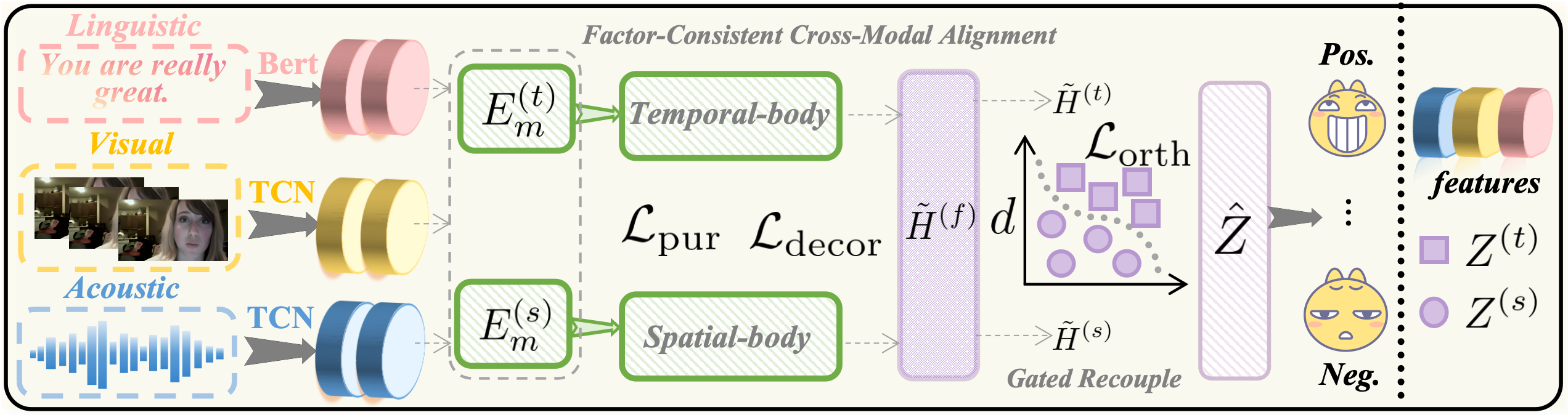}
    \caption{Architecture illustration of TSDA, which decouples temporal and spatial factors, aligns them with FCCA, and adaptively fuses them through the Gated Recouple Module.}
    \label{fig:TSDA}
\end{figure*}

\section{Introduction}

Multimodal Sentiment Analysis (MSA) integrates Linguistic, Visual, and Auditory evidence to infer sentiment, and supports applications in human computer interaction, assistive agents, and media understanding~\cite{rts-vit,JMT,d2r,mmin}. Compared with unimodal learning, multimodal models can exploit complementary cues across sources, but representation learning remains difficult due to heterogeneity in feature distributions and signal reliability across modalities~\cite{self-mm,DEVA,cta-net,MGJR}.

In recent years, researchers have proposed numerous multimodal models to address this issue. Prior works follows two main directions. The first direction learns cross modal interaction in a shared latent space. Such as, Decoupled Modality Distillation transfers knowledge from a strong modality to a weak one while decoupling teacher and student objectives to stabilize training under imbalance~\cite{dmd}. Expert based frameworks allocate modality specific experts and learn their aggregation to balance specialization and coverage~\cite{EMOE}. Contrastive Eigendecomposition introduces a contrastive objective that separates common and exclusive signals, which sharpens decision boundaries during fusion~\cite{confede}. The second direction reduces distribution gaps by factorizing each modality into invariant and specific subspaces while retaining idiosyncratic cues. MISA aligns modalities in a shared invariant subspace and enforces orthogonality in private subspaces to provide complementary views for downstream fusion~\cite{misa}. Subsequent disentanglement based methods tighten the invariant alignment and promote diversity in private spaces using disparity or contrastive regularization~\cite{FDMER,FDRL}. Language centric variants decompose biased textual features to improve robustness and transfer across domains~\cite{DLF,MCIS}.

Despite steady progress, most pipelines still encode each modality as a single mixed embedding before interaction, forcing temporal dynamics and spatial or structural context to be learned jointly. For sentiment this design is fragile: decisive evidence often resides in sparse temporal patterns such as prosodic bursts, hesitations, and micro expressions, whereas static context such as identity, timbre, and sentence level polarity is abundant and tends to dominate optimization~\cite{sctnet}. When these cues are entangled, the objective is largely explained by high variance static components, which attenuates the temporal trajectories that determine polarity and leads to brittle predictions~\cite{Semi-IIN,CGGM}. We refer to this failure as spatiotemporal information asymmetry and diagnose it with perturbations that selectively disrupt one factor while preserving the other, namely temporal shuffle and static swap. These observations motivate a formulation that treats temporal and spatial factors explicitly and prevents their mixing prior to fusion.

Hence, we address this challenge with TSDA, Temporal Spatial Decouple before Act. For each modality, a temporal encoder maps ordered tokens or frames into a temporal sequence and a spatial encoder produces a time agnostic structural set. Interaction is then restricted to factor consistent paths through Factor Consistent Cross Modal Alignment, which uses masked cross attention with a block diagonal pattern so that temporal features align only with temporal counterparts across modalities and spatial features align only with spatial counterparts. Residual cross factor leakage is suppressed by factor purity supervision at the token level and decorrelation regularization at the summary level, which preserves complementarity while preventing contamination. A gated recouple module then recombines the aligned temporal and spatial summaries on a per instance basis. An orthogonality regularizer keeps the two streams well conditioned during fusion. 
Our main contributions as follows:
\begin{itemize}
\item We propose \textbf{TSDA}, a novelty \textbf{Temporal-Spatial Decouple before Act} framework that decouples dynamics and structure before interaction, eliminating pre-fusion coupling and alleviating spatiotemporal asymmetry.

\item We develop \textbf{Factor-Consistent Cross-Modal Alignment (FCCA)}, which aligns temporal and spatial streams independently via block-diagonal masked attention. Factor-purity supervision and decorrelation further ensure separation while preserving complementarity.

\item We propose a \textbf{Gated Recouple Module (GR)} that integrates temporal and spatial summaries with instance-wise calibrated weights, guided by factor disagreement and confidence signals, interventional supervision, and an orthogonality regularizer to mitigate static dominance.

\end{itemize}

\section{Method}
\subsection{Task Formulation}
Let $\mathcal{M}=\{L,V,A\}$ denote language, visual, and acoustic modalities. For each modality $m\in\mathcal{M}$, we observe a token sequence $X_m=\{x_{m,1},\ldots,x_{m,T_m}\}$ of length $T_m$. The goal is to predict a sentiment target $y$ for classification or regression. Each $X_m$ carries short-lived temporal cues and relatively stable spatial or structural context. Conventional pipelines embed these factors jointly before cross-modal interaction, which biases learning toward static components. As shown in Fig.~\ref{fig:TSDA}, we instead separate time and space at the representation stage prior to any cross-modal exchange.

\begin{table*}[htbp]
\setlength{\tabcolsep}{6pt}
\renewcommand{\arraystretch}{1.0}
\caption{Performance comparison on the benchmarks, results under aligned/unaligned settings as $a / b$ ($a$: aligned, $b$: unaligned).}
\begin{tabular}{l||cccc||cccc}
\toprule
\multirow{2}{*}{\textbf{Methods}}
& \multicolumn{4}{c}{\textbf{CMU-MOSI}}
& \multicolumn{4}{c}{\textbf{CMU-MOSEI}} \\
& MAE ($\downarrow$) & ACC$_7$ (\%) & ACC$_2$ (\%) & F1 (\%)
& MAE ($\downarrow$) & ACC$_7$ (\%) & ACC$_2$ (\%) & F1 (\%) \\
\midrule
\midrule
LMF~\cite{LMF}          & 0.931 / 0.963 & 36.9 / 31.1 & 78.7 / 79.1 & 78.7 / 79.1
                        & 0.564 / 0.565 & 52.3 / 51.9 & 84.7 / 83.8 & 84.5 / 83.9 \\
MuLT~\cite{MuLT}        & 0.936 / 0.933 & 35.1 / 33.2 & 80.0 / 80.3 & 80.1 / 80.3
                        & 0.572 / 0.556 & 52.3 / 53.2 & 82.7 / 84.0 & 82.8 / 84.0 \\
TFN~\cite{TFN}          & 0.953 / 0.995 & 31.9 / 35.3 & 78.8 / 76.5 & 78.9 / 76.6 & 0.574 / 0.573 & 50.9 / 50.2 & 80.4 / 84.2 & 80.7 / 84.0 \\
MISA~\cite{misa}        & 0.754 / 0.742 & 41.8 / 43.6 & 84.2 / 83.8 & 84.2 / 83.9
                        & 0.543 / 0.557 & 52.3 / 51.0 & 85.3 / 84.8 & 85.1 / 84.8 \\
FDMER~\cite{FDMER}      & -     / 0.725 & -   / 44.2  & -    / 84.6 & -    / 84.7
                        & -     / 0.536 & -   / 53.8  & -    / 84.1 & -    / 84.0 \\
ConFEDE~\cite{confede}  & -     / 0.742 & -   / 46.3  & -    / 84.2 & -    / 84.2
                        & -     / 0.523 & -   / 54.9  & -    / 81.8 & -    / 82.3 \\
Self-MM~\cite{self-mm}  & 0.738 / 0.724 & 45.3 / 45.7 & 84.9 / 83.4 & 84.9 / 83.6
                        & 0.540 / 0.535 & 53.2 / 52.9 & 84.5 / 85.3 & 84.3 / 84.8 \\
MMIN~\cite{mmin}        & - / 0.741     & - / -       & 83.5 / 85.5 & 83.5 / 85.51
                        & - / 0.542     & - / -       & 83.8 / 85.9 & 83.9 / 85.76 \\
DMD~\cite{dmd}          & 0.721 / 0.721 & 46.2 / 46.7 & 83.2 / 84.0 & 83.2 / 84.0
                        & 0.546 / 0.536 & 52.4 / 53.1 & 84.8 / 84.7 & 84.7 / 84.7 \\
DEVA~\cite{DEVA}        & -     / 0.730 & -   / 46.3  & -    / 84.4 & -    / 84.5
                        & -     / 0.541 & -   / 52.3  & -    / 83.3 & -    / 82.9 \\
DLF~\cite{DLF}          & -     / 0.731 & -   / 47.1  & -    / 85.1 & -    / 85.1
                        & -     / 0.536 & -   / 53.9  & -    / 84.4 & -    / 85.3 \\
EMOE~\cite{EMOE}        & 0.710 / 0.697 & 47.7 / 47.8 & 85.4 / 85.4 & 85.4 / 85.3
                        & 0.536 / 0.533 & 54.1 / 53.9 & 85.3 / 85.5 & 85.3 / 85.5 \\
\textbf{TSDA (Ours)}    & \textbf{0.695} / \textbf{0.680} & \textbf{48.6} / \textbf{48.5} & \textbf{86.3} / \textbf{86.5} & \textbf{86.2} / \textbf{86.5}
                        & \textbf{0.529} / \textbf{0.527} & \textbf{54.9} / \textbf{54.9} & \textbf{86.3} / \textbf{86.4} & \textbf{86.2} / \textbf{86.5} \\
\bottomrule
\end{tabular}
\label{tab:main}
\end{table*}

\subsection{Temporal and Spatial Disentanglement}
Temporal models capture long-range dependencies, but a single mixed embedding blends dynamic cues with static context and biases learning. We address this at the source by mapping each input into two factor-specific bodies that target complementary aspects: temporal evolution and spatial structure. For each modality $m$, a temporal encoder $E^{(t)}_m$ produces a sequence of temporal tokens and a spatial encoder $E^{(s)}_m$ yields time-invariant structural tokens:
\begin{equation}
F_m^{(t)}=E^{(t)}_m(X_m)\in\mathbb{R}^{T_m\times d_t},\quad
F_m^{(s)}=E^{(s)}_m(X_m)\in\mathbb{R}^{S_m\times d_s}.
\end{equation}
Here $T_m$ and $S_m$ are the numbers of temporal and spatial tokens, and $d_t$ and $d_s$ are the embedding widths. The spatial granularity $S_m$ depends on the modality, for example patches or regions for video, segments for audio, and sentence or phrase tokens for language. In practice, $E^{(t)}_m$ and $E^{(s)}_m$ are lightweight heads on unimodal backbones: the temporal head preserves order along the sequence axis, while the spatial head aggregates local evidence into stable structural tokens. This factorization prepares the factor-consistent alignment that follows and prevents early co-adaptation between dynamics and context.

\subsection{Factor-Consistent Cross-Modal Alignment}
After disentanglement, each modality $m\in\{L,V,A\}$ yields a temporal sequence $F_m^{(t)}\in\mathbb{R}^{T_m\times d_t}$ and a spatial set $F_m^{(s)}\in\mathbb{R}^{S_m\times d_s}$. Conventional fusion mixes factors in one space, which induces cross-factor interference and static dominance. FCCA aligns like with like by running two independent cross-modal attentions with a block-diagonal message passing structure. This gives a formal guarantee of factor consistency. Let $H^{(t)}=\mathrm{concat}(\{F_m^{(t)}\}_m)$ and $H^{(s)}=\mathrm{concat}(\{F_m^{(s)}\}_m)$. For factor $f\in\{t,s\}$ we compute masked attention and summaries
\begin{equation}
\tilde{H}^{(f)}=\mathrm{softmax}\!\left(\frac{Q^{(f)}{K^{(f)}}^{\top}}{\sqrt{d_f}}+\log M^{(f)}\right)V^{(f)}
\end{equation}

\begin{equation}
Z^{(f)}=\mathrm{Pool}\big(\tilde{H}^{(f)}\big),
\end{equation}

where $Q^{(f)},K^{(f)},V^{(f)}$ are linear projections of $H^{(f)}$, and $M^{(f)}$ is a binary mask that permits attention only within the same factor. The mask renders the operator block-diagonal with respect to the factor index, so no temporal-to-spatial messages are exchanged during alignment. Finally, we get the representation information $Z^{(f)}$.

Residual leakage is controlled at two levels. At the token level, a discriminator $D:\mathbb{R}^{d_f}\to[0,1]$ predicts whether an aligned token originates from the temporal stream, and we optimize the purity objective:
\begin{equation}
\mathcal{L}_{\mathrm{pur}}
=\mathbb{E}\!\left[-\log D(\tilde{h}^{(t)})\right]
+\mathbb{E}\!\left[-\log\big(1-D(\tilde{h}^{(s)})\big)\right].
\end{equation}
At the summary level, we limit overlap between factors by penalizing second-order and nonlinear dependence:
\begin{equation}
\mathcal{L}_{\mathrm{decorr}}
=\lambda_c\,\cos^2\!\big(Z^{(t)},Z^{(s)}\big)
+\lambda_h\,\mathrm{HSIC}\!\big(Z^{(t)},Z^{(s)}\big).
\end{equation}
In combination, the block-diagonal attention supplies the structural constraint, purity supervision enforces factor identifiability, and decorrelation reduces redundancy between summaries, which together preserve complementarity while preventing cross-factor contamination.

\subsection{Gated Recouple}
We recombine the aligned temporal and spatial summaries with instance-wise weights that reflect reliability. A plain scalar gate tends to overweight static context when temporal cues are sparse, so we calibrate the gate using signals from FCCA and interventional supervision. Let $Z^{(t)}$ and $Z^{(s)}$ be the aligned summaries; define disagreement $d=1-\cos\!\big(Z^{(t)},Z^{(s)}\big)$ and factor confidences $c_t=\mathrm{mean}\!\big(D(\tilde{H}^{(t)})\big)$ and $c_s=\mathrm{mean}\!\big(1-D(\tilde{H}^{(s)})\big)$. The gate takes $\phi=[Z^{(t)};Z^{(s)};d;c_t;c_s]$ and outputs $g=\sigma(w^{\top}\phi+b)\in(0,1)$. We form $\hat{Z}=g\,U_t Z^{(t)}+(1-g)\,U_s Z^{(s)}$ with projection heads $U_t\in\mathbb{R}^{d\times d_t}$ and $U_s\in\mathbb{R}^{d\times d_s}$ and impose an orthogonality regularizer $\mathcal{L}_{\mathrm{orth}}=\|U_t^{\top}U_s\|_F^2$ to reduce collinearity and improve gate identifiability. Gate calibration uses two interventions: temporal shuffle that destroys order while preserving static structure, and static swap that replaces the spatial summary while retaining temporal trajectories; the gate is penalized if it assigns high weight to the corrupted factor, and a soft prior encourages larger $g$ when $(c_t-c_s)$ and $d$ are larger.

\subsection{Training Objective}
Let $\mathcal{L}_{\mathrm{task}}$ be cross-entropy for classification or mean-squared error for regression. The overall loss is
\begin{equation}
\label{eq:loss-total}
\mathcal{L}
=\mathcal{L}_{\mathrm{task}}
+\alpha\,\mathcal{L}_{\mathrm{pur}}
+\beta\,\mathcal{L}_{\mathrm{decorr}}
+\gamma\,\mathcal{L}_{\mathrm{orth}},
\end{equation}
with nonnegative weights $\alpha,\beta,\gamma$. The objective ensures factor-consistent cross-modal interaction, factor purity at the token level, and complementary yet decorrelated temporal and spatial summaries.

\section{Experiments}
\subsection{Experimental Settings}
\textbf{Datasets.} We evaluate TSDA on standard benchmarks. CMU-MOSI~\cite{Cmu-mosi}: 2{,}199 segments with aligned text, audio, and vision; split 1{,}284/229/686. CMU-MOSEI~\cite{Cmu-mosei}: 23{,}453 clips from 1{,}000 speakers; split 16{,}326/1{,}871/4{,}659. Sentiment labels range from $-3$ to $+3$. \textbf{Metrics.} We report ACC-2, ACC-7, macro-F1, and MAE, following EMOE~\cite{EMOE}. \textbf{Implementation.} Models are implemented in PyTorch and trained on NVIDIA A100 GPUs. We use Adam with weight decay $1{\times}10^{-5}$, batch size $16$, and up to $50$ epochs with early stopping. Five-fold cross-validation is employed.

\subsection{Comparison with State-of-the-art Methods}

Table~\ref{tab:main} presents results on CMU-MOSI and CMU-MOSEI under both aligned and unaligned settings. TSDA achieves the best performance on all metrics across the two datasets. On MOSI, TSDA surpasses the strongest baseline EMOE by reducing MAE from 0.710/0.697 to 0.695/0.680, and by improving ACC$_7$, ACC$_2$, and F1 by about 1\% on average. On MOSEI, TSDA achieves the lowest MAE of 0.529/0.527 and the highest accuracy and F1, consistently outperforming prior disentanglement- and fusion-based methods. These improvements are attributable to the temporal–spatial disentanglement that prevents static-dominance bias, the factor-consistent alignment that enforces clean temporal and spatial interactions, and the gated recoupling that adaptively balances reliability across factors. Together, these mechanisms yield more faithful sentiment representations and stable gains across both aligned and unaligned conditions, highlighting the robustness and effectiveness of TSDA relative to existing approaches.

\begin{table}[t]
\centering
\caption{Ablation studies of TSDA on the benchmarks.}
\setlength{\tabcolsep}{3pt}
\renewcommand{\arraystretch}{0.9}
\begin{tabular}{l||cc||cc}
\toprule
\multirow{2}{*}{Models}
& \multicolumn{2}{c||}{CMU-MOSI}
& \multicolumn{2}{c}{CMU-MOSEI} \\
& MAE ($\downarrow$) & ACC$_7$ (\%) & MAE ($\downarrow$) & ACC$_7$ (\%) \\
\midrule
\midrule
\textbf{TSDA (Ours)}   & \textbf{0.680} & \textbf{48.5} & \textbf{0.527} & \textbf{54.9} \\
\midrule
\multicolumn{5}{c}{\emph{Importance of Modality}} \\
w/o L. & 1.015 & 35.2 & 0.835 & 39.6 \\
w/o A. & 0.818 & 44.8 & 0.624 & 51.0 \\
w/o V. & 0.842 & 44.4 & 0.647 & 50.4 \\
\midrule
\multicolumn{5}{c}{\emph{Importance of Representations}} \\
w/o Temporal       & 0.726 & 46.0 & 0.552 & 52.5 \\
w/o Spatial        & 0.716 & 46.8 & 0.546 & 53.0 \\
w/o ST Disen.      & 0.731 & 45.7 & 0.555 & 52.2 \\
w/o FCCA           & 0.728 & 45.5 & 0.552 & 51.9 \\
\midrule
\multicolumn{5}{c}{\emph{Different Fusion Mechanisms}} \\
Sum                & 0.711 & 47.8 & 0.537 & 54.0 \\
Concat             & 0.714 & 47.5 & 0.538 & 53.9 \\
GR (Ours)          & 0.703 & 47.9 & 0.536 & 54.1 \\
\midrule
\multicolumn{5}{c}{\emph{Importance of Regularization}} \\
w/o $\mathcal{L}_{\mathrm{pur}}$    & 0.722 & 46.5 & 0.548 & 52.9 \\
w/o $\mathcal{L}_{\mathrm{decorr}}$ & 0.713 & 46.9 & 0.541 & 53.3 \\
w/o $\mathcal{L}_{\mathrm{orth}}$   & 0.714 & 47.1 & 0.542 & 53.4 \\
CE Loss                               & 0.736 & 45.7 & 0.556 & 52.0 \\
\bottomrule
\end{tabular}
\label{tab:ablation-tsda}
\end{table}

\subsection{Ablation Studies}

We conduct ablation experiments to assess the necessity of each component in TSDA. Results on MOSI and MOSEI are reported in Table~\ref{tab:ablation-tsda}, covering modality contributions, representation disentanglement, fusion strategies, and regularization designs. 

\textbf{Importance of Modality.} Excluding any modality causes substantial performance loss. Removing language leads to the largest degradation, confirming that lexical polarity anchors sentiment decisions. Excluding audio or vision results in smaller but still notable drops, showing that these streams provide complementary cues for robust multimodal inference. 

\textbf{Importance of Representations.} Removing the temporal stream produces a sharper decline than removing the spatial stream, consistent with the hypothesis that sentiment is often carried by short-lived prosodic and visual dynamics. Eliminating temporal–spatial disentanglement further reduces accuracy, indicating that factor separation itself is critical. Even with disentanglement retained, removing FCCA harms performance, which demonstrates that explicit factor-consistent alignment is essential to prevent cross-factor interference and static dominance. 

\textbf{Different Fusion Mechanisms.} Simple sum or concatenation fusion lags behind, as these operations cannot adaptively balance streams. The proposed gated recoupling improves results by weighting factors based on reliability, but still underperforms the full model. This suggests that adaptive reweighting alone is insufficient without the structural constraints of FCCA, and that both mechanisms are jointly required. 

\textbf{Importance of Regularization.} Removing factor purity supervision $\mathcal{L}_{\mathrm{pur}}$ leads to the largest degradation on both MOSI and MOSEI, showing its central role in preserving clean temporal factors. Removing decorrelation $\mathcal{L}_{\mathrm{decorr}}$ or orthogonality $\mathcal{L}_{\mathrm{orth}}$ also weakens performance, though the impact is milder and comparable across the two datasets. Training solely with cross-entropy yields the worst results overall, which underscores that purity, decorrelation, and orthogonality are complementary components that together enforce factor separation and stability.

These ablations validate the design motivation of TSDA, disentanglement supplies distinct temporal and spatial bodies, FCCA guarantees consistent alignment, and gated recoupling adaptively integrates them. Together, they resolve spatiotemporal asymmetry and deliver stable improvements across benchmarks.

\subsection{Visualization Analysis}
\begin{figure}[htbp]
\centerline{\includegraphics[width=0.85\columnwidth]{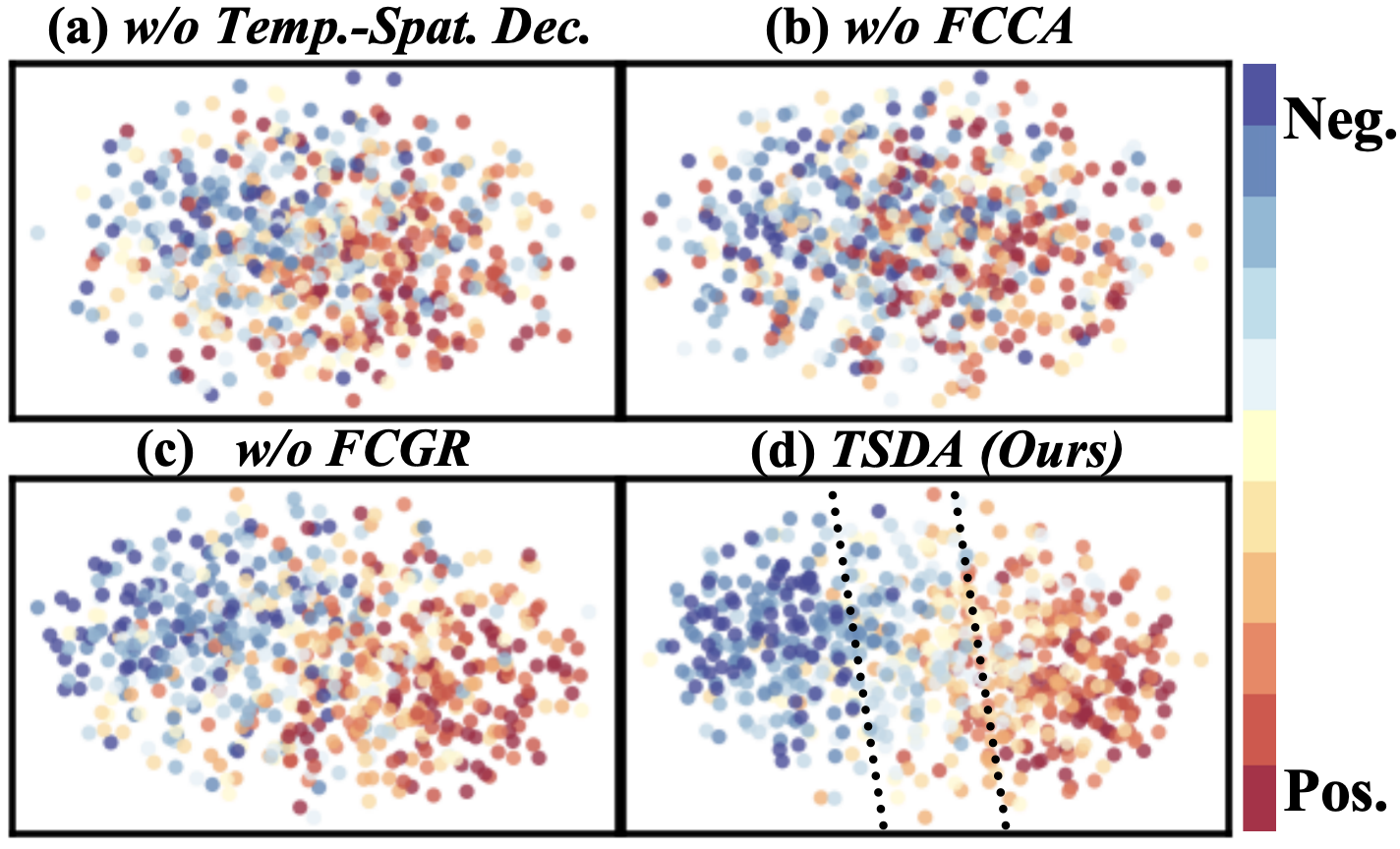}}
\caption{T-SNE visualization on MOSI. Red indicates stronger positive sentiment. TSDA yields the best structure.}
\label{fig:Qualitative-Analysis}
\end{figure}

\textbf{Qualitative analysis.} To prove the effectiveness of the TSDA, we visualize MOSI test embeddings with t-SNE to inspect the feature distribution (Fig.~\ref{fig:Qualitative-Analysis}). w/o Temp.-Spat. Dec., points are diffuse, polarities overlap, and intensity lacks ordering. Removing FCCA after disentanglement yields partial gains, yet temporal and spatial streams still mix and static cues dominate. Adding GR sharpens the polarity continuum and suppresses isolated clusters by down-weighting unreliable factors, though minor cross-factor leakage persists. The full TSDA produces the most compact, monotonic gradient aligned with affect intensity, which is the desired geometry for regression. The progression indicates that disentanglement is necessary, FCCA prevents cross-factor interference, and GR calibrates per-sample reliability.

\begin{figure}[htbp]
\centerline{\includegraphics[width=\columnwidth]{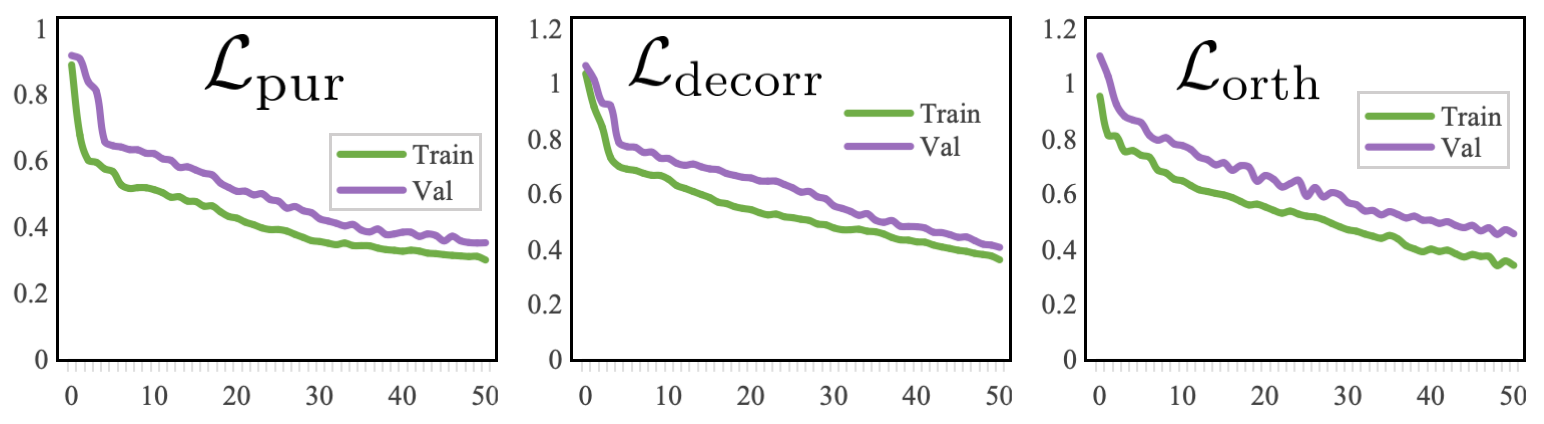}}
\caption{Regularization curves during MOSI training. Similar trends are observed on MOSEI.}
\label{fig:Regularization}
\end{figure}

\textbf{Regularization Trend}. During MOSI and MOSEI training, the constraints $\mathcal{L}_{\mathrm{pur}}$, $\mathcal{L}_{\mathrm{decorr}}$, and $mathcal{L}_{\mathrm{orth}}$ are almost monotonically decreasing and converge smoothly. As shown in Fig.~\ref{fig:Regularization}, the factor-purity classifier for token identification improves steadily and then stabilizes, indicating that FCCA learns factor-distinct representations. In parallel, the cosine dependence between summaries $\cos^{2}\!\big(Z^{(t)},Z^{(s)}\big)$ drops in step with $\mathcal{L}_{\mathrm{decorr}}$, and the loss $\mathcal{L}_{\mathrm{orth}}$ descends without oscillation. These trends show that TSDA enforces factor separation, limits cross factor leakage, and maintains stable optimization toward the end task.

\begin{figure}[htbp]
\centerline{\includegraphics[width=\columnwidth]{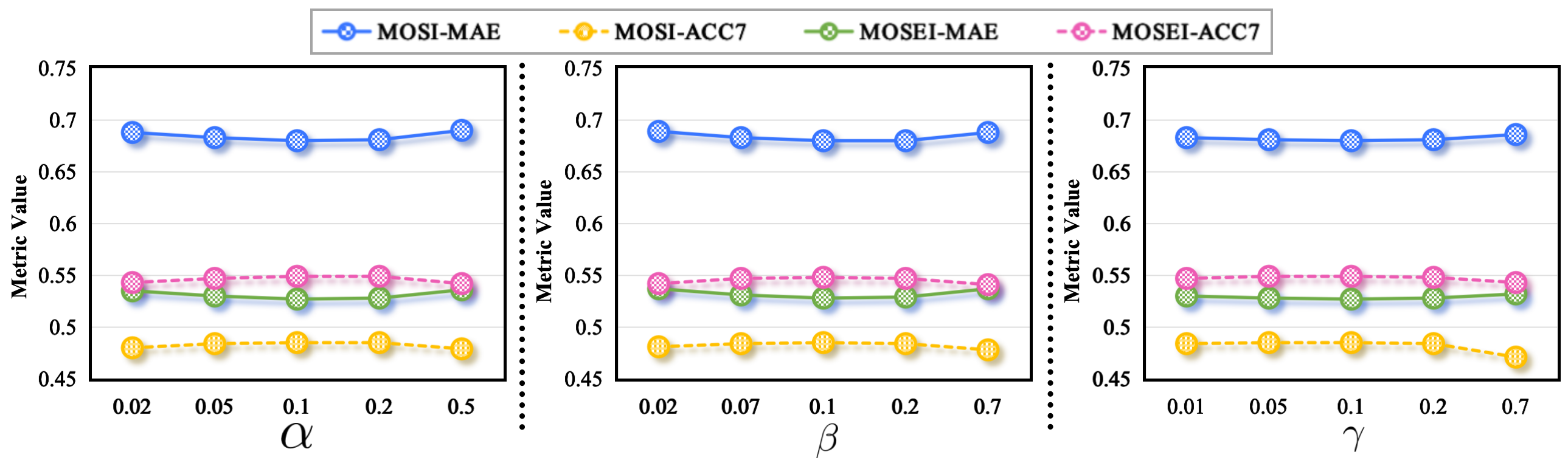}}
\caption{Sensitivity of TSDA to $\alpha$, $\beta$, and $\gamma$ on benchmarks. Performance remains stable across values.}
\label{fig:sensitivity}
\end{figure}

\textbf{Sensitivity Analysis.} We assess robustness by sweeping the loss weights in Eq.~\eqref{eq:loss-total}. For each hyperparameter $\alpha$, $\beta$, and $\gamma$, we vary its value over a range while fixing the others to the defaults used in the main experiments, and evaluate on MOSI and MOSEI under both aligned and unaligned settings. Fig.~\ref{fig:sensitivity} reports MAE and $\mathrm{ACC}_7$ as the hyperparameters change. The curves exhibit only marginal fluctuations across all configurations, indicating that TSDA is insensitive to the precise choices of $\alpha$, $\beta$, and $\gamma$, and that the observed gains arise from the architecture rather than careful tuning.

\section{Conclusion}
In this paper, we introduced TSDA, a novelty Temporal-Spatial Decouple before Act framework that addresses spatiotemporal heterogeneity in multimodal sentiment analysis by disentangling temporal dynamics and spatial context prior to interaction. The Factor-Consistent Cross-Modal Alignment ensures reliable alignment across factors, while the Gated Recouple adaptively balances their contributions. Extensive experiments demonstrate that this design effectively mitigates spatiotemporal asymmetry and delivers consistent improvements on benchmark datasets. In future work, we plan to extend TSDA to the field of human-computer interaction.

\bibliographystyle{IEEEbib}
\bibliography{strings,refs}
\end{document}